\documentclass[11pt,a4paper]{article}
\usepackage[hyperref]{acl2020}
\usepackage{multirow}
\usepackage{mathtools}

\usepackage{times}
\usepackage{latexsym}

\usepackage{microtype}

\usepackage{verbatim}
\usepackage[disable]{todonotes}

\newcommand{\archna}[2][]
{\todo[color=green,inline, #1]{\textbf{Archna:} #2}}

\aclfinalcopy 

\title{The Panacea Threat Intelligence and Active Defense Platform}

\author{
Adam Dalton,\textsuperscript{\rm 1}
Ehsan Aghaei,\textsuperscript{\rm 3}
Ehab Al-Shaer,\textsuperscript{\rm 3}
Archna Bhatia,\textsuperscript{\rm 1} \\
\textbf{Esteban Castillo,\textsuperscript{\rm 4}
Zhuo Cheng,\textsuperscript{\rm 3}
Sreekar Dhaduvai,\textsuperscript{\rm 2}
Qi Duan,\textsuperscript{\rm 3}
Md Mazharul Islam,\textsuperscript{\rm 3}}\\ 
\textbf{
Younes Karimi,\textsuperscript{\rm 2}
Amir Masoumzadeh,\textsuperscript{\rm 2}
Brodie Mather,\textsuperscript{\rm 1}
Sashank Santhanam,\textsuperscript{\rm 3}
}\\
\textbf{
Samira Shaikh,\textsuperscript{\rm 3}
Tomek Strzalkowski\textsuperscript{\rm 4}
Bonnie J. Dorr\textsuperscript{\rm 1}
}\\
\textsuperscript{\rm 1}IHMC, Ocala, FL, USA, \{adalton,abhatia,bmather,bdorr\}@ihmc.us\\ 
\textsuperscript{\rm 2}SUNY, Albany, NY, USA \{sdhaduvai,amasoumzadeh,ykarimi\}@albany.edu\\ 
\textsuperscript{\rm 3}UNCC, Charlotte, NC, USA \{eaghaei,ealshaer,zcheng5,qduan,\\mislam7,ssantha1,sshaikh2\}@uncc.edu\\ 
\textsuperscript{\rm 4}Rensselaer Polytechnic Institute, Troy, NY, USA \{castie2,tomek\}@rpi.edu
}

\date{}

\begin{document}
\maketitle

\begin{abstract}
\end{abstract}
We describe Panacea, a system that supports natural language processing (NLP) components for active defenses against social engineering attacks. We deploy a pipeline of human language technology, including Ask and Framing Detection, Named Entity Recognition, Dialogue Engineering, and Stylometry. Panacea processes modern message formats through a plug-in architecture to accommodate innovative approaches for message analysis, knowledge representation and dialogue generation. The novelty of the Panacea system is that uses NLP for cyber defense and engages the attacker using bots to elicit evidence to attribute to the attacker and to waste the attacker's time and resources. 

\section{Introduction}
Panacea (Personalized AutoNomous Agents Countering Social Engineering Attacks) actively defends against social engineering (SE) attacks. \textit{Active} defense refers to engaging an adversary during an attack to extract and link attributable information while also wasting their time and resources in addition to preventing the attacker from achieving their goals. This contrasts with \textit{passive} defenses, which decrease likelihood and impact of an attack \cite{denning2014framework} but do not engage the adversary. 

SE attacks are formidable because intelligent adversaries exploit technical vulnerabilities to avoid social defenses, and social vulnerabilities to avoid technical defenses \cite{hadnagy2015phishing}. A system must be socially aware to find attack patterns and indicators that span the socio-technical space. Panacea approaches this by incorporating the F3EAD (Find, Fix, Finish, Exploit, Analyze, and Disseminate) threat intelligence cycle \cite{gomez2011targeting}.
The \textit{find} phase identifies threats using language-based and 
message security approaches.
The \textit{fix} phase gathers relevant and necessary information to engage the adversaries and plan the mitigations that will prevent them from accomplishing their malicious goals.
The \textit{finish} phase 
performs a decisive and responsive action
in preparation for the \textit{exploit} phase for future attack detection.
The \textit{analysis} phase exploits intelligence from conversations with the adversaries and places it in a persistent knowledge base where it can be linked to other objects and studied additional context. 
The \textit{disseminate} phase makes this intelligence 
available to all components to improve performance in subsequent attacks.

\begin{figure}[h]
\includegraphics[width=\linewidth]{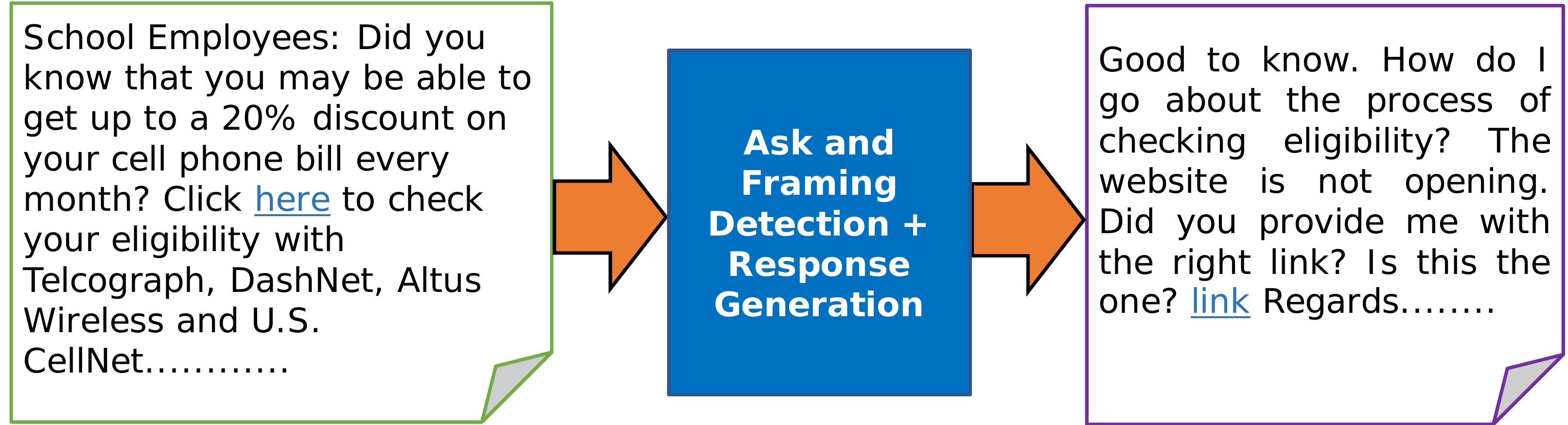}
\caption{Active Defense against Social Engineering: Attacker's email (left) yields bot's response (right)}
\label{fig:ask-framing-response}
\end{figure}

Panacea's value comes from NLP capabilities for cyber defense coupled with end-to-end plug-ins for ease of running NLP over real-world conversations. Figure~\ref{fig:ask-framing-response} illustrates Panacea's active defense in the form of conversational engagement, diverting the attacker while also delivering a link that will enable the attacker's identity to be unveiled. 

 \vspace*{-.2in}
\subsection{Use Cases}

Panacea's primary use cases are: (1) monitoring a user's inbox to detect SE attacks; and (2) 
engaging the attacker to gain attributable information about their true identity while preventing attacks from succeeding. 
Active SE defense tightly integrates offensive and defensive capabilities to detect and respond to SE campaigns. 
Engaging the adversary 
uniquely enables extraction of indicators required to confidently classify a communication as malicious. Active defenses also carry significant risk because
engagement
can potentially harm an
individual's or organization's reputation.
Thus, high confidence classification is vital. 
 
\subsubsection{Monitoring and Detection}

Panacea includes an initial protection layer based on the analysis of incoming messages. Conceptual users include end users and IT security professionals.
Each message 
is processed and assigned a label of friend, foe, or unknown,
taking into account headers and textual information of each message. The data obtained from this analysis is converted into threat intelligence and stored in a knowledge graph for use in subsequent phases, e.g.,
for meta analysis and 
message analysis in a broader context
within a thread or 
in similar messages delivered to multiple users.

\subsubsection{Engagement and Attribution}
Passive defenses
are finished once a threat is discovered, defused, and deconstructed;
at this point Panacea's active defenses become engaged. 
Panacea's active defenses 
respond to the attacker's demands, reducing the risk that the attacker will catch on that they've been fingered. As such,
any requests made by 
Panacea are more likely to be fulfilled by the attacker, bulwarked by hopes of eventual payoff. Such requests are implemented as a collection of flag seeking strategies built on top of a conversational theory of \textit{asks}. Flags are
collected using information extraction techniques. 
Future work includes inferential logic and deception detection to unmask an attacker and separate them from feigned identities used to gain trust.

\section{Related Work}

Security in online communication is a challenge due to: (1) attacker's speed outpacing that of defenders to maintain indicators \cite{Zhang2006PhindingPE}; (2) phishing site quality high enough that users ignore alerts \cite{Egelman:2008:YWE:1357054.1357219}; (3) user training falling short as users forget material and fall prey to previously studied attacks \cite{caputo2013going}; the divergent goals of the attacker and defender \cite{Li2020EndtoEndTN}; and (4) defensive system maintainers who may ignore account context, motivations, and socio-economic status of the targeted user \cite{Oliveira2017-phishing}. Prior studies \cite{bak2008,kar2006} 
demonstrate human susceptibility to SE attacks. Moving from bots that detect such attacks to those that produce ``natural sounding'' responses, i.e., conversational agents that engage the attacker to elicit identifying information, is the next advance in this arena. 

Prior work extracts information from email interactions \cite{Dada2019}, applies
supervised learning to identify email signatures and forwarded messages \cite{Carvalho2004},
and classifies email content 
into different structural sections 
\cite{Lampert2009}.
Statistical and rule-based heuristics 
extract users' names and aliases \cite{Yin2011} and
structured script representations 
determine whether an email resembles a password reset email typically sent from an organization's IT department \cite{li-goldwasser-2019-encoding}. 
Analysis of
chatbot responses \cite{prakhar-2019-open-domain-dialogue} yields human-judgement correlation improvements.
Approaches above differ from ours in that they require extensive model training.

Our approach relates to work on conversational agents, e.g., response generation using neural models \cite{gao2019neural,santhanam2019survey}, topic models  \cite{dziri2018augmenting}, self-disclosure for targeted responses \cite{RavichanderB18}, topic models \cite{Bhakta:2015}, and other NLP analysis \cite{Yuki:2016}. All such approaches are limited to a pre-defined set of topics, constrained by the training corpus. 
Other prior work focuses on persuasion detection/prediction \cite{mckeown:aaai-2018} but for judging when a persuasive attempt might be successful, whereas
Panacea aims to achieve effective dialogue for countering (rather than adopting) persuasive attempts. 
Text-based semantic analysis is also used for SE detection \cite{kimcatch}, but not for \textit{engaging} with an attacker. Whereas a bot might be employed to warn a potential victim that an attack is underway, our bots communicate with a social engineer in ways that elicit identifying information.

Panacea's architecture is inspired by state-of-the-art systems in cyber threat intelligence. MISP \cite{wagner2016misp} focuses on information sharing from a 
community of trusted organizations. MITRE's Collaborative Research Into Threats (CRITs) \cite{mitrecrits} platform is, like Panacea, built on top of the Structured Threat Intelligence eXchange (STIX) specification.
Panacea differs from these in that it is part of operational active defenses, rather than solely an analytical tool for incident response and threat reporting. 

\section{System Overview}

Panacea's processing workflow is inspired by Stanford's CoreNLP annotator pipeline \cite{manning-EtAl:2014:P14-5}, but with a focus on using NLP to power active defenses against SE. A F3EAD-inspired phased analysis and engagement cycle is employed to conduct active defense operations. 
The cycle is triggered when a message arrives and is deconstructed into STIX threat intelligence objects. 
Object instances for the identities of the sender and all recipients are found or created in the knowledge base. Labeled relationships are created between those identity objects and the message itself. 

Once a message is ingested, plug-in components process the message in the \textit{find} phase, yielding a response as a 
JSON object that is used
by plug-in components 
in subsequent phases. Analyses performed in this phase include message part decomposition,
named entity recognition, and email header analysis. The \textit{fix} phase 
uses components dubbed \textit{deciders}, which perform a meta-analysis of the results from the \textit{find} phase to determine if and what type of an attack is taking place. 
\textit{Ask detection} provides a fix on what the attacker is going after in the \textit{fix} phase, if an attack is indicated. 
Detecting an attack advances the cycle to the \textit{finish} phase, 
where response generation is activated.

Each time Panacea successfully elicits a response from the attacker, the new message is \textit{exploited} for attributable information, such as the geographical location of the attack and what organizational affiliations they may have. This information is stored as structured intelligence in the knowledge base which triggers the \textit{analysis} phase, wherein the threat is re-analyzed in a broader context. Finally, Panacea disseminates threat intelligence so that humans can build additional tools and capabilities to combat future threats. \archna{please make sure descriptions of all the phases are consistent at different places they are mentioned in the paper.}


\section{Under the Hood}

\begin{figure}[!h]
\includegraphics[height=8cm, width=\linewidth]{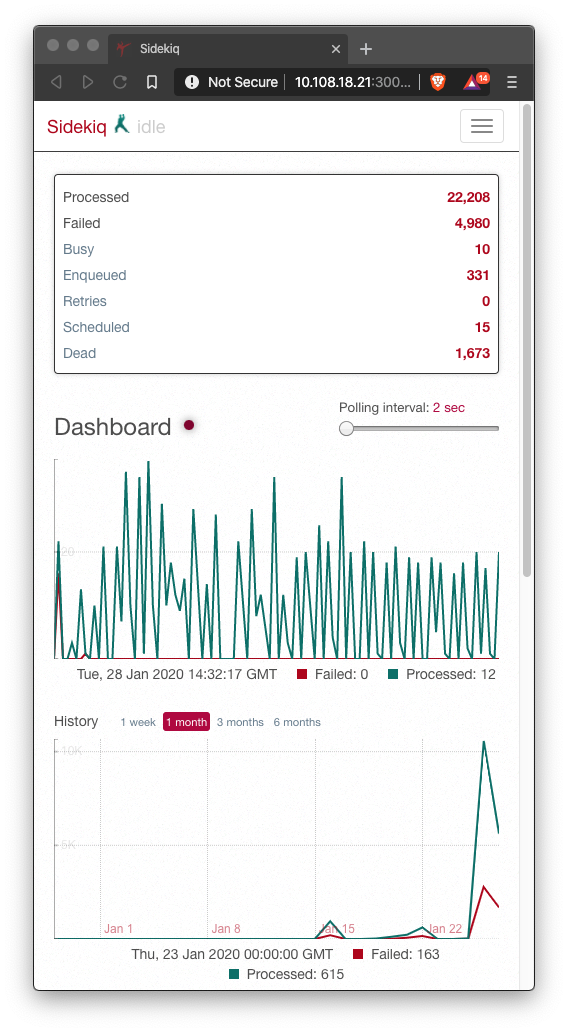}
\caption{Panacea components run asynchronously in the background
for scaling and so new components can be added and removed based on the underlying task.}
\label{fig:sidekiq}
\end{figure}

Panacea's main components are presented: (1) Message Analysis Component; and (2) Dialogue Component. The resulting system is capable of handling the thousands of messages a day that would be expected in a modern organization, including failure recovery and scheduling jobs for the future. Figure \ref{fig:sidekiq} shows Panacea throughput while operating over a one month backlog of emails, SMS texts, and LinkedIn messages.\archna{Figure \ref{fig:sidekiq} does not show what proportion was emails, what proportion SMS and LinkedIn respectively. Unless only one of them was a big majority and the other two types of messages were rare, I think it would look good that Panacea was able to handle messages coming from different sources and use them without getting overwhelmed.}

\subsection{Message Analysis Component}
Below we describe the structural aspects of messages and their associated processing.

\subsubsection{Email Header Classification}
When communication takes place over a network, metadata is extracted that serves as a user fingerprint and a source for reputation scoring. Email headers, for example, contain authentication details and information about the mail servers that send, receive, and relay messages as they move from outbox to inbox. To distinguish between benign and malicious emails, Panacea applies a multistage email spoofing, spamming, and phishing detector consisting of: (1) a signature-based detector, (2) an active investigation detector, (3) a receiver-oriented anomaly detector, and (4) a sender-oriented anomaly detector.

\subsubsection{Email Content Classification}
Dissecting email headers is not enough for detecting malicious messages. Many suspicious elements are related to email bodies that contain user messages related to a specific topic and domain. Analyzing email content provides valuable insight for detecting threats in conversations and a solid understanding of the content itself. Panacea incorporates machine learning algorithms that, alongside of  header classifiers, digest email exchanges:

\paragraph{Benign/non-benign classifier:} Word embedding vectors \cite{Bengio2006,Mikolov2013} trained on email samples from different companies (e.g., Enron) are extracted using neural networks \cite{Sherstinsky2018}, i.e., back-propagation model with average word vectors as features. This classifier provides a binary prediction regarding the nature of emails (friend or foe).

\paragraph{Email threat type classifier:} Spam, phishing, malware, social-engineering and propaganda are detected, providing fine-grained information about the content of emails and support for motive detection (i.e., attacker's intention).

\paragraph{Email zone classifier:} Greetings, body, and signature are extracted using word embedding implemented as recurrent neural network with handcrafted rules, thus yielding senders, receivers and relevant entities to enable response generation.

All classifiers support active detection of malicious emails and help in the engagement process of automated bots. Additionally, all trained models have an overall accuracy of 90\% using a cross validation approach against well known email collections like Enron \cite{Klimt2004} and APWG \cite{Oest2018} among other non-public datasets, which makes them reasonably reliable in the context of passive defenses.   

\subsubsection{Behavioral Modeling}

If an adversary is able to compromise a legitimate account, then the header and content classifiers will not be sufficient to detect an attack. The social engineer is able to  extract contacts of the account owner and send malicious content on their behalf, taking advantage of the reputation and social relationships attributed to the hijacked account. Two distinctive approaches address these issues:

\paragraph{Impersonation Detector:} Sender entities are extracted from the email message and a personalized profile is created for each one, with communication habits, stylometric features, and social network. The unique profiled model is used to assess whether this email has been written and sent by an account's legitimate owner. If a message arrives from a sender that does not have a profile, Panacea applies similarity measures to find other email addresses for the unknown entity. This serves as a defense against impersonation attacks where the social engineer creates an email account using a name and address similar to the user of an institutional account for which a model is available. If Panacea links the unknown account to an institutional account, then that account's model is used to determine whether a legitimate actor is using an unknown account, or a nefarious actor is attempting to masquerade as an insider in order to take advantage of the access such an account would have.

\textbf{Receiving Behavior Classifier.} Individual profiles are built for the receiving behavior of each entity (how and with whom this entity communicates) and new emails are evaluated against the constructed models. To build unique profiles, all messages sent to each particular entity are collected.

\subsubsection{Deciders}

Panacea must have high confidence in determining that a message is coming from an attacker before deploying active defense mechanisms. A strategy-pattern approach fits different meta-classifiers to different situations. Four classification strategies, called \textit{Deciders}, combine all component analyses after a message is delivered to an inbox to make the final \textit{friend/foe} determination. The Decider API expects all component analyses to include a \textit{friend/foe} credibility score using six levels defined by the Admiralty Code \cite{JDP:2-00}. Deciders may be deterministic through the application of rule based decision making strategies or they may be trained to learn to identify threats based on historical data.

\subsubsection{Threat Intelligence}

Panacea stores component analysis results in a threat intelligence knowledge  base for aggregation of 
attack campaigns with multiple turns, targets, and threads.
The  knowledge base adheres to STIX 2.0 specifications and implements MITRE's ATT\&CK framework \cite{strom2017finding} to enable attribution and anticipatory mitigations of sophisticated SE attacks. Panacea recognizes indicators of compromise based on features of individual emails as well as historical behavior of senders and recipients. Intrusion sets and campaigns are thus constructed when malicious messages \archna{do we have gold truth on malicious messages, may be we should specify that explicitly, but may be it is already  mentioned, I might have missed it.} are discovered subsequently linked to threat actors based on attribution patterns, such as IP address, message templates, socio-behavioral indicators, and linguistic signatures. This feature set was prioritized 
to work with Unit 42's ATT\&CK Playbook Viewer.
The knowledge base uses a PostgreSQL database backend with an application layer built with Ruby on Rails.

\subsection{Dialogue Component}

Panacea's dialogue component consists of three key sub-components: \textit{Ask/Framing Detection} (to determine the attacker's demand), \textit{Motive Detection} (to determine the attacker's goal), and \textbf{Response Generation} (to reply to suspicious messages).

\subsubsection{Ask/Framing Detection}

\label{ask-framing}
Once an email is processed as described above, linguistic knowledge and structural knowledge are used to extract candidate Ask/Framing pairs and to provide the final confidence-ranked output.

\paragraph{Application of Linguistic Knowledge:}
Linguistic knowledge is employed to detect both the \textit{ask} (e.g., buy gift card) and the \textit{framing} (e.g., lose your job, get a 20\% discount). An ask 
may be, for example, a request for something (GIVE) or an action (PERFORM). On the other hand, framing may be a reward (GAIN) or a risk (LOSE), for example. Ask/framing detection relies on Stanford CoreNLP constituency parses and dependency trees \cite{CoreNLP:2018}, coupled with \textit{semantic role labeling} (SRL) \cite{AllenNLP:2017}, to identify the main action and arguments. For example, \textit{click \underline{here}} yields \textit{click} as the \textit{ask} and its argument \textit{here}. 

Additional constraints are imposed through the use of a lexicon based on Lexical Conceptual Structure (LCS) \cite{Dor:18a,dorr-voss:2018}, derived from
a pool of team members' collected suspected scam/impersonation emails. Verbs from these emails were 
grouped as follows:
\begin{tabular}{l}
$\bullet$~PERFORM: connect, copy, refer\\
$\bullet$~GIVE: administer, contribute, donate\\
$\bullet$~LOSE: deny, forget, surrender\\
$\bullet$~GAIN: accept, earn, grab, win
\end{tabular}

Additional linguistic processing includes: (1) categorial variation \cite{Nizar:2003} to map between different parts of speech, e.g., \textit{reference}(N) $\rightarrow$ \textit{refer}(V) enables detection of an explicit ask from \textit{you can reference your gift card}; and (2) verbal processing to eliminate spurious asks containing verb forms such as \textit{sent} or \textit{signing} in \textit{sent you this email because you are signing up}. 

\paragraph{Application of Structural Knowledge:} Beyond meta-data processing described previously, the email body is further pre-processed before linguistic elements are analyzed.  Lines are split where \texttt{div},  \texttt{p},  \texttt{br}, or  \texttt{ul} tags are encountered. Placeholders are inserted for hyperlinks. Image tags are replaced with their alt text. All styling, scripting, quoting, replying, and signature are removed.

Social engineers employ different link positionings to present ``click bait,''
e.g., ``Click {\color{blue}\underline{here}}'' or ``Contact me ({\color{blue}\underline{jw11@example.com}}).''
Basic link processing assigns the link to the appropriate ask (e.g., \textit{click here}). Advanced link processing ties together an email address with its corresponding PERFORM ask (e.g., \textit{contact me}), even if separated by intervening material.

\paragraph{Confidence Score and Top Ask:} Confidence scores are heuristically assigned:  (1) Past tense events are assigned low or 0 confidence;  (2) The vast majority of asks associated with URLs (e.g., {\color{blue}\underline{jw11@example.com}}) are found to be PERFORM asks with highest confidence (0.9); (3) a GIVE ask combined with any ask category (e.g., \textit{contribute \$50}) is less frequently found to be an ask, thus assigned slightly lower confidence (0.75); and  (4) GIVE by itself is even less likely found to be an ask, thus assigned a confidence of 0.6 (e.g., \textit{donate often}). Top ask selection then selects highest confidence asks at the aggregate level of a single email. This is crucial for downstream processing, i.e., response generation in the dialogue component. For example, the ask ``PERFORM contact ({\color{blue}\underline{jw11@example.com}})'' is returned as the top ask for ``\textit{Contact me. ({\color{blue}\underline{jw11@example.com}})}.''

\subsubsection{Motive Detection}
\label{motive}

In addition to the use of distinct tools for detecting linguistic knowledge, Panacea extracts the attacker's intention, or \textit{motive}. Leveraging the attacker's demands (asks), goals (framings) and message attack types (from the threat type classifier), the Motive Detection module maps to a range of possible motive labels: \textit{financial information}, \textit{acquire personal information}, \textit{install malware}, \textit{annoy recipient}, etc. Motive detection maps to such labels from top asks/framings and their corresponding threat types. Examples are shown here:
\hspace*{-.35in}
$$\footnotesize{ \underbrace{\mathit{Give}}_\text{ \footnotesize Ask} + \underbrace{\mathit{Finance\:info}}_\text{\footnotesize Ask type} + \underbrace{\mathit{Spam}}_\text{\footnotesize Email threat} \rightarrow \textit{Financial\:info}} $$
\hspace*{-.35in}
$$\footnotesize{\underbrace{\mathit{Gain}}_\text{ \footnotesize Framing} + \underbrace{\mathit{Credentials}}_\text{\footnotesize Ask type} + \underbrace{\mathit{Malware}}_\text{\footnotesize Email threat} \rightarrow \textit{Install\:malware}}$$

These motives are used later for enhancing a response generation process which ultimately creates automatic replies for all malicious messages detected in the Panacea platform.

\subsubsection{Response Generation}
\label{rgcomponent}

Response generation is undertaken by a bot using templatic approach to yield appropriate responses based on a hierarchical attack ontological structure and ask/framing components. The hierarchical ontology contains 13 major categories (e.g., \textit{financial details}). Responses focus on wasting the attacker's time or trying to gain information from the attacker while moving along \textit{F3EAD} threat intelligence cycle \cite{gomez2011targeting} to ensure that the attacker is kept engaged. The response generation focuses on the \textit{find, finish and exploit} states. The bot goes after name, organization, location, social media handles, financial information, and is also capable of sending out malicious links that obtain pieces of information about the attacker's computer. 

A dialogue state manager decides between time wasting and information seeking based on motive, ontological structure and associated ask/framing of the message. For example, if an attack message has motive \textit{financial details} and ontological structure of \textit{bank information}, coupled with a PERFORM ask, the dialogue state manager moves into an information gathering phase and produces this response: ``\textit{Can you give me the banking information for transferring money? I would need the bank name, account number and the routing information. This would enable me to act swiftly.}'' On the other hand if the attacker is still after financial information but not a particular piece of information, the bot wastes time, keeping the attacker in the loop.


\section{Evaluation}

Friend/foe detection (Message Analysis) and response generation (Dialogue) are evaluated for effectiveness of Panacea as an effective intermediary between attackers and potential victims.

\subsection{Message Analysis Module}

The DARPA ASED program evaluation tests header and content modules against messages for friend/foe determination. Multiple sub-evaluations check system accuracy in distinguishing malicious messages from benign ones, reducing the false alarm rate, and transmitting appropriate messages to dialogue components for further analysis. Evaluated components yield $\sim$90\% accuracy.
Components adapted for detecting borderline exchanges (\textit{unknown} cases) are shown to help dialogue components request more information for potentially malicious messages.

\subsection{Dialogue Module}

The ASED program evaluation also tests the dialogue component. Independent evaluators communicate with the system without knowledge of whether they are interacting with humans or bots. Their task 
is to engage in a dialogue for as many turns as necessary. Panacea bots are able to sustain conversations for an average of 5 turns (across 15 distinct threads). Scoring applied by independent evaluators yield a rating of 1.9 for their ability to display human-like communication (on a scale of 1\textendash3; 1$=$bot, 3$=$human). This score is the highest amongst all other competing approaches (four other teams) in this independent program evaluation.

\section{Conclusions and Future Work}
Panacea is an operational system that processes communication data into actionable intelligence and provides active defense capabilities to combat SE. 
The F3EAD active defense cycle was chosen because it fits the SE problem domain, but specific phases 
could be changed to address different problems. For example, a system using the Panacea processing pipeline could ingest academic papers on a disease, process them with components designed to extract biological mechanisms, then engage with 
paper authors 
to ask clarifying questions and search for additional literature to review, while populating a knowledge base containing the critical intelligence for the disease of interest. 


Going forward, the plan is to improve Panacea's plug-in infrastructure so that it is easier to add capability without updating Panacea itself. This is currently possible as long as new components use the same REST API as existing components. 
The obvious next step is to formalize Panacea's API. We have found value to
leaving it open at this early state of development as we discover new challenges and solutions to problems that emerge 
in building a large scale system focused on the dangers and opportunities in human language communication.

\section*{Acknowledgments}
This work was supported by DARPA through AFRL Contract
FA8650-18- C-7881 and Army Contract
W31P4Q-17-C-0066. All statements of fact, opinion or
conclusions contained herein are those of the authors and
should not be construed as representing the official views
or policies of DARPA, AFRL, Army, or the U.S. Government.

\bibliography{acl2020,aaai2020}
\bibliographystyle{acl_natbib}

\end{document}